\documentclass[twoside,11pt]{article}

\usepackage{jmlr2e}

\graphicspath{{./img/}}
\usepackage{amsmath,graphicx}
\usepackage{algorithm}
\usepackage[]{algpseudocode}

\jmlrheading{1}{2015}{1-48}{4/00}{10/00}{Amedeo Buonanno and Francesco A.N. Palmieri}

\ShortHeadings{Towards Building Deep Networks  with Bayesian Factor Graphs}{Buonanno and Palmieri}
\firstpageno{1}

\begin{document}

\title{Towards Building Deep Networks  with \\ Bayesian Factor Graphs}

\author{\name Amedeo Buonanno \email amedeo.buonanno@unina2.it \\
       \addr Dipartimento di Ingegneria Industriale e dell'Informazione\\
       Seconda Universit\`a di Napoli (SUN)\\
       via Roma 29, Aversa (CE), Italy
       \AND
       \name Francesco A.N.\ Palmieri \email francesco.palmieri@unina2.it \\
       \addr Dipartimento di Ingegneria Industriale e dell'Informazione\\
       Seconda Universit\`a di Napoli (SUN)\\
       via Roma 29, Aversa (CE), Italy}

\editor{}

\maketitle

\begin{abstract}
We propose a Multi-Layer Network based on the Bayesian framework of the Factor Graphs in Reduced Normal Form (FGrn) applied to a two-dimensional lattice. 
The Latent Variable Model (LVM) is the basic building block of a quadtree hierarchy built on top of a bottom layer of random variables that represent pixels of an image, a feature map, or more generally a collection of spatially distributed discrete variables.
The multi-layer architecture implements a hierarchical data representation that, via belief propagation,  can be used for learning and inference. Typical uses are pattern completion, correction and classification.
The FGrn paradigm provides great flexibility and modularity and appears as a promising candidate for building deep networks: the system can be easily extended by introducing new and different (in cardinality and in type) variables. Prior knowledge, or supervised information, can be introduced at  different scales. The FGrn paradigm provides a handy way for  building all kinds of architectures by interconnecting only three types of units: Single Input Single Output (SISO) blocks, Sources and Replicators.  The network is designed like a circuit diagram and the belief messages flow bidirectionally in the whole system. The learning algorithms operate only  locally within each block. The framework is demonstrated in this paper in a three-layer structure applied to images extracted from a standard data set.
\end{abstract}


\section{Introduction}

Building efficient  representations for images, and more in general for sensory data, is one of the central issues in signal processing. The problem  has received  much attention in the literature of the last thirty years because, almost invariably, the extraction of information from observations requires that  raw data is translated first into ``feature maps" before classification or filtering.  

Recent striking results with ``deep networks" have generated  much attention in machine learning on what is known as {\em Representation Learning} (see \citep{Bengio2012} for a review). The main idea of these methods is to learn multiple representation levels as progressive abstractions of the input data.    
The creation of a feature hierarchy permits to the structure inside the data to emerge at different scales combining more and more complex features as we go upward in the hierarchy \citep{Bengio2011}, \citep{Bengio2014}. 

For image understanding this process is somewhat biologically plausible too. There is a vast literature that postulates the hierarchical organization of the primary visual cortex (V1). The neurons become selective for stimuli that are increasingly complex, from simple oriented bars and edges to moderately complex features, such as a combination of orientations, to complex objects \citep{Serre2010}. We do not derive our models from the biology, but we cannot avoid recognizing that the most successful artificial systems paradigms  share some common features with what is observed in nature.   

In building an artificial system, one of the key issues is to provide sufficiently general methods that can be applied across different kinds of sensory data, letting learning capture most of the specificity of the application context. This is why in this work we focus on a Bayesian network approach, that has the advantage of being totally  general with respect to the type of data processed defining a framework that can easily fuse information coming from different sensor modalities. In a Bayesian network the information flow is bi-directional via belief propagation and can easily accommodate various kinds of inferences  for pattern completion, correction and classification.   

Various architectures have been proposed as adaptive Bayesian graphs \citep{Koller2009}, \citep{Barber2012}, but in our case the use of Factor Graphs \citep{Forney2001}, \citep{Loeliger2004},  specially in the simplified Reduced Normal Form \citep{Palmieri2013}, allows better modularity. Message propagation follows  standard sum-product rules, but the system is built as the interconnection of only SISO blocks, source blocks and replicators with learning equations defined in a totally localized fashion.  

In this paper we proposes a new deep architecture based on FGrn applied to a two-dimensional lattice. The Latent Variable Model (LVM) \citep{Murphy2012}, \citep{Bishop1999}, also known as Autoclass \citep{Cheeseman1996} is the basic building block of a quadtree hierarchy. Learning is totally  localized inside the SISO blocks that constitute the LVMs. The complete system can be seen as a partitioned  type of Latent Tree Model \citep{MouradZhang2013}. 

The application of the Bayesian model to images shows how the hierarchy extracts the primitives at various scales and how, via bi-directional belief propagation, it provides a reliable structure for learning and inference in various modes.   

In  Section 2 we review some of the related literature while 
in Section 3 we introduce notations and the basics of belief propagation in FGrn.
In Section 4 we present the LVM, i.e.  the building block for the multi-layer architecture that is presented  in Section 5  with the learning strategy and the Encoding/Decoding process.
In Section 6 we apply learning and inference to images from a standard data set.
Section 7 includes conclusive remarks and suggestions for further work.

\section{Related Work}

The vast literature on the deep representation learning (see the extensive overview in \citep{Schmidhuber2014}) can be mostly divided in two main lines of research: the first one is based on probabilistic graphical models such as the Restricted Boltzmann Machine (RBM) \citep{Hinton2006}, \citep{HintonSalakhutdinov2006b}, \citep{Lee2008} and the second one is based on neural network models as the autoencoder \citep{Bengio2007}, \citep{Ranzato2006}. At the same time several unsupervised feature learning algorithms have been proposed: Sparse Coding \citep{Olshausen1996},\citep{Lee2008}, RBM \citep{Hinton2006}, Autoencoders \citep{Bengio2007}, \citep{Ranzato2006}, \citep{Vincent2008}, K-means \citep{Coates2012}. Other models based on the memory-prediction theory of brain have also been proposed \citep{Hawkins2004}, \citep{Dileep2008}.

Confining our interest to probabilistic graphical models, the most natural choice for modeling the spatial interactions between pixels (or patches) in the image is a two-dimensional lattice (Markov Random Field - MRF) where the nodes represent the pixels (or patches) and the potential functions are associated to the edges between adjacent nodes \citep{Wainwright08}. Various tasks in image processing such as denoising, segmentation, and super-resolution, can be treated as an inference problem on the MRF. For these models convergence of the inference is not guaranteed and even if for large-scale models it is intractable, approximate and sub-optimal methods have been often used: Markov Chain Monte Carlo methods \citep{GemanGeman1984}, \citep{Gelfand1990}, variational methods \citep{Jordan99}, \citep{Beal2003}, graph cut \citep{Boykov99} and Belief Propagation \citep{Xiong2007}.

An alternative strategy to MRF is to replace the 2D lattice with a simpler and approximate model as multiscale (or multiresolution) structures like quadtrees. These have the advantages of allowing the application of efficient tree algorithms to perform exact inference with the trade off that the model is imperfect \citep{Luettgen1993}, \citep{Bouman94}, \citep{Nowak1999}, \citep{Laferte2000}, \citep{Willsky2002}.
Another problem of the quadtree structure is the non locality  since  two neighboring pixels may or may not share a common parent node depending on their position on the grid. For avoiding this problem Wolf et al. have proposed a Markov cube adding additional connections at the different levels \citep{Wolf2010}.

On the quadtree structure  inference can be performed using the {\em belief propagation} algorithm that  was originally proposed for inferences on trees where exact solutions are guaranteed \citep{Pearl1988}. When the graph has loops, open issues still remain about the accuracy of inferences, even though often the bare application of standard belief propagation may already provide satisfactory results ({\em loopy belief propagation}) \citep{Yedidia2005}, \citep{Frean2008}.  When the problem can be reduced to a tree, belief propagation provides exact marginalization and algorithms for learning latent trees have been proposed \citep{Choi2011} with  successful applications to computer vision.

A very appealing approach to directed Bayesian graphs for visualization and manipulation, that has not found its full way in the applications, is the Factor Graph (FG) representation and in particular the so-called Normal Form (FGn) \citep{Forney2001}, \citep{Loeliger2004}. This formulation is very appealing because it provides an easy way to visualize and manipulate Bayesian graphs - much like in block diagrams. Factor Graphs assign variables to edges and functions to nodes. Furthermore, in the Reduced Normal Form (FGrn), through the use of replicator units (or equal constraints), the graph is reduced to an architecture in which each variable is connected to two factors at most \citep{Palmieri2013}. In this way any architecture (deep or shallow) can be built as the interconnection of only three types of units: Single Input Single Output (SISO) blocks, Sources and Diverters (Replicators), with the learning equations defined locally (Figure \ref{fig:intro}).

This is the framework on which this paper is focused because the designed network resembles a circuit diagram with belief messages more easily visualized as they flow into SISO blocks and travel through replicator nodes \citep{BuoWirn2014}. This paradigm provides extensive modularity because replicators act like buses and can be used as expansion nodes when we need to augment an existing model with new variables.
Parameter learning, in this representation, can be approached in a unified way because we can concentrate on a unique rule for training any SISO, or Source, factor-block in the system, regardless of  its location (visible or hidden). 

In our previous work \citep{Palmieri_cip2014} we have reported some preliminary results on a multi-layer convolution Bayesian Factor Graph built as a stack of HMM-like trees. Each layer is built from a latent model trained on the messages coming from the layer below. The structure is loopy, but our  experience has shown that BP performs well in recovering information from the deep parts of the network: the upper layers contain progressively larger-scale information that is pipelined to the bottom for pattern completion or correction across sequences.  

In this work we step back and confine our attention to a quadtree structure, for which no loops are present and inference is exact.  We have found that this framework, even if just a tree, has great potential of being used in a very large number of applications for its inherent modularity at the expenses of a certain growth in computational complexity. The complexity issue will be discussed in the paper.
To our knowledge the FGrn framework has never been used to build deep networks.

\section{Factor Graphs in Reduced Normal Form}
\label{sec:not}
\begin{figure} [!]
\centering
\includegraphics[width=0.6\linewidth]{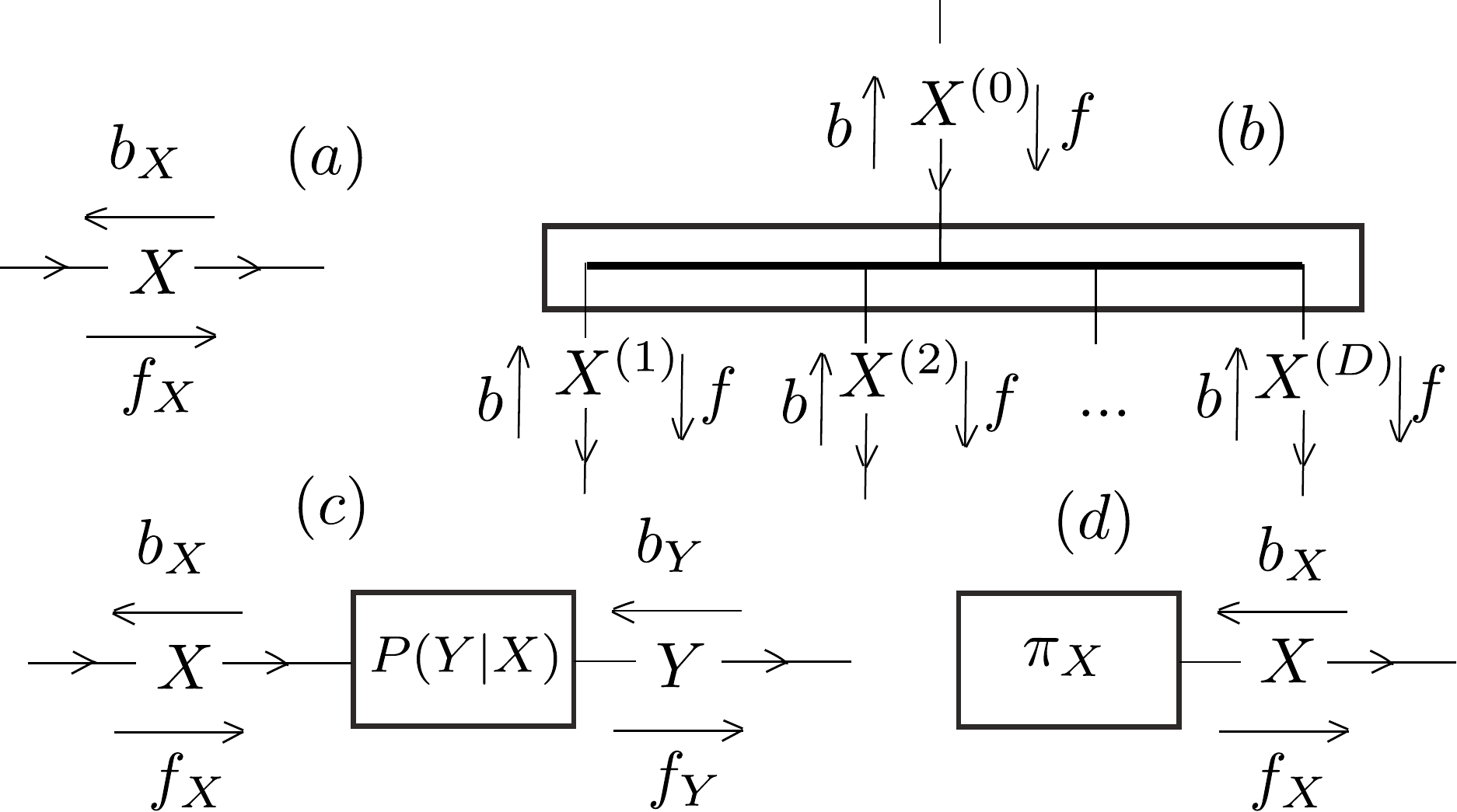}
\caption{FGrn components: (a) a variable branch; (b) a diverter; (c) a SISO block; (d) a source block.}
\label{fig:intro}
\end{figure}

In the FGrn framework \citep{Palmieri2013} the Bayesian graph  is reduced to a simplified form composed only by {\em Variables}, {\em Replicators} (or {\em Diverters}), {\em Single-Input/Single-Output (SISO) blocks} and {\em Source blocks}. Even though various architectures have been proposed in the literature for Bayesian graphs \citep{Loeliger2004}, we have found that the FGrn framework is much easier to handle, it is more suitable to define unique learning equations \citep{Palmieri2013} and it is more suited for distributed implementations. The blocks needed to compose any architecture are shown in Figure  \ref{fig:intro}. In our notation we avoid the upper arrows for the messages and assign a direction to each variable branch for unambiguous definition of forward and backward messages. 

\noindent

For a {\em variable} $X$ (Figure \ref{fig:intro}(a)) that takes values in the discrete alphabet
${\cal X}= \{\xi_1, \xi_2, \ldots,$ $ \xi_{d_X} \}$, forward and backward messages are in function form $b_X(\xi_i)$ and $f_X(\xi_i)$, $i=1:d_X$ and in vector form ${\bf b}_X=(b_X(\xi_1), b_X(\xi_2),\hdots,b_X(\xi_{d_X}))^T$ and ${\bf f}_X=(f_X(\xi_1), f_X(\xi_2),\hdots,f_X(\xi_{d_X}))^T$.  
All messages are proportional ($\propto$) to discrete distributions and may be normalized to sum to one.

Comprehensive knowledge about $X$ is contained in the posterior distribution $p_X$ obtained through the product rule, $p_X(\xi_i) \propto f_X(\xi_i) b_X(\xi_i)$, $i=1:d_X$, in function form, or ${\bf p}_X \propto {\bf f}_X \odot {\bf b}_X$, in vector form, where $\odot$ denotes the element-by-element product. The result of each product is proportional to a distribution and can be normalized to sum one (it is a good practice to keep messages normalized to avoid poorly conditioned products).   

The {\em replicator} (or diverter) (Figure \ref{fig:intro}(b)) represents the equality constraint with the variable $X$ replicated ($D + 1$) times. Messages for incoming and outgoing branches carry different forward and backward information. Messages that leave the  block are obtained as the product of the incoming ones:
$b_{X^{(0)}}(\xi_i) \propto \prod_{j=1}^{D}  b_{X^{(j)}}(\xi_i)$; $f_{X^{(k)}}(\xi_i) \propto f_{X^{(0)}}(\xi_i) \prod_{j=1, j \neq k}^{D}  b_{X^{(j)}}(\xi_i)$, $k=1:D$, $i=1:d_x$ in function form. 
In vector form: ${\bf b}_X^{(0)} \propto \odot_{j=1}^{D}  {\bf b}_X^{(j)}$; ${\bf f}_X^{(k)} \propto {\bf f}_X^{(0)} \odot_{j=1, j \neq k}^{D}  {\bf b}_X^{(j)}$, $k=1:D$.

The {\em SISO block} (Figure \ref{fig:intro}(c)) represents the conditional probability matrix of $Y$ given $X$. More specifically if $X$ takes values in the discrete alphabet ${\cal X}=\{\xi_1,\xi_2,...,\xi_{d_X} \}$ and $Y$ in  ${\cal Y}=\{\upsilon_1,\upsilon_2,...,\upsilon_{d_Y} \}$, $P(Y|X)$ is the 
$d_X \times d_Y$ row-stochastic matrix $P(Y|X)=[Pr\{ Y=\upsilon_j | 
X=\xi_i\}]_{i=1:d_X}^{j=1:d_Y}=[\theta_{ij}]_{i=1:d_X}^{j=1:d_Y}$. Outgoing messages are:  $f_Y(\upsilon_j) \propto \sum_{i=1}^{d_X} \theta_{ij} f_X(\xi_i)$;  $b_X(\xi_i) \propto \sum_{j=1}^{d_Y} \theta_{ij} b_Y(\upsilon_j)$, in function form. In vector form: ${\bf f}_Y \propto P(Y|X)^T {\bf f}_X$;  ${\bf b}_X \propto P(Y|X) {\bf b}_Y$. 

The {\em source block} in Figure \ref{fig:intro}(d) is the termination for the independent source variable $X$. More specifically $\pi_X$ is the $d_X$-dimensional prior distribution on $X$ with the outgoing message $f_X(\xi_i)=\pi_X(\xi_i), i = 1 : d_X$ in function form, or 
${\bf f}_X={\boldsymbol \pi}_X$ in vector form. The backward message ${\bf b}_X$ coming from the network can be combined with the forward ${\bf f}_X$ for final posterior on  $X$.
   
For the reader not familiar with the factor graph framework, it should be emphasized that the above rules are rigorous translation of Bayes' theorem and marginalization. For a more detailed  review, refer to our recent works \citep{Palmieri2013}, \citep{BuoWirn2014} (or to the classical papers \citep{Loeliger2004} \citep{Kschischang2001}).

Parameters (probabilities) in the SISO  and the source blocks must be learned from examples solely on the backward and forward flows available locally. 
We set the learning problem as an EM algorithm to maximize global likelihood \citep{Palmieri2013}. Focusing on a specific SISO block, if all the other network parameters have been fixed, maximization of global likelihood translates in the local problem from examples $({\bf f}_{X[n]}, {\bf b}_{Y[n]})$, $n=1,...,N_e$ 
\begin{equation}
\left\{ \begin{array}{l}
\min_{\theta} ~~ - \sum_{n=1}^{N_e}  \log \left( {\bf f}_{X[n]}^T~ \theta~ {\bf b}_{Y[n]} \right), \\
\theta ~~~~ {\rm row-stochastic}.
\end{array} \right. 
\label{eq:mlp}
\end{equation}
After adding a stabilizing term to the cost function and applying KKT conditions we obtain the following algorithm \citep{Palmieri2013}.

\begin{algorithm}
\caption{Learning Algorithm for SISO block}\label{LearningAlgo}
\begin{algorithmic}[1]
\Procedure{Learning Algo}{}
\State Initialize $\theta$ to uniform rows: $\theta=(1/d_Y) {\bf 1}_{d_X \times d_Y}$

\For{$i=1:d_X$}
\State $ftmp(i) = \sum_{n=1}^{N_e} f_{X[n]}(i)$
\EndFor

\For{$it = 1:N_{it}$}
\For{$n=1:N_e$}
\State $den(n) = {\bf f}_{X[n]}^T \theta {\bf b}_{Y[n]}$
\EndFor

\For{$i=1:d_X$}
\For{$j=1:d_Y$}
\State $tmpSum = \sum_{n=1}^{N_e}  \frac{f_{X[n]}(i) b_{Y[n]}(j)}{den(n)}$

\State $\theta_{ij} \gets \frac{\theta_{ij}}{ftmp(i)} \cdot tmpSum$,


\EndFor
\EndFor

\State Row-normalize $\theta$
\EndFor
\EndProcedure

\noindent
{\footnotesize 
(We have used the shortened notation $f_{X[n]}(\xi_i)=f_{X[n]}(i)$,  $b_{Y[n]}(v_j)=b_{Y[n]}(j)$).}

\end{algorithmic}
\end{algorithm}
  
In Algorithm \ref{LearningAlgo} there are  three main blocks and the complexity in the worst case is $O(N_e \cdot d_X \cdot d_Y \cdot N_{it})$.
The algorithm is a fast multiplicative update with no free parameters. The iterations usually converge in a few steps and the presence of the normalizing factors  makes the algorithm numerically very stable. The algorithm has been discussed and compared to other similar updates in  \citep{Palmieri2013}. 

The updates for the source block are immediate if we set the forward messages
${\bf f}_{X[n]}$ to a uniform distribution and consider any row of $\theta$ to be the target distribution.   

\section{Bayesian Clustering}

\begin{figure}[!]
\centerline{\includegraphics[width=0.8\linewidth]{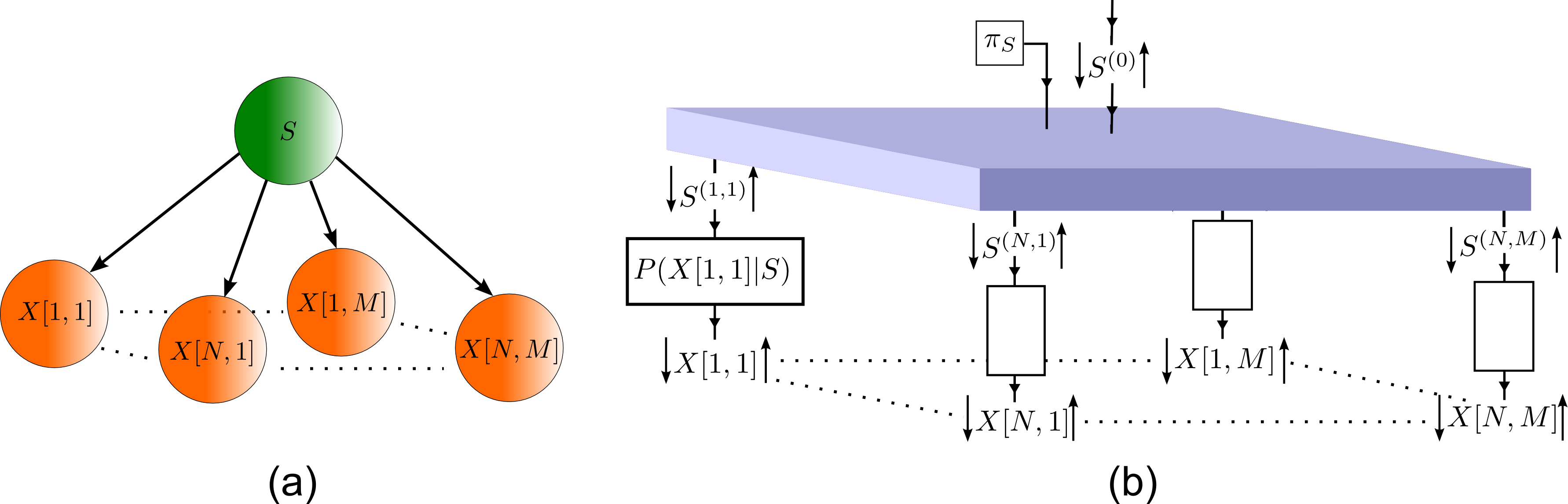}}
\caption{A ($N \cdot M$) - tuple with the Latent Variable as a Bayesian graph (left) and as a Factor Graph in Reduced Normal Form (right). Only the first SISO block has been explicitly described. }
\label{fig:lat1}
\end{figure}

For the architectures that will follow, the basic building block is the {\em Latent-Variable Model} (LVM) shown in Figure \ref{fig:lat1}. 
At the bottom of each LVM there are $N \cdot M$  variables  
$X[n,m]$, $n=1:N$, $m=1:M$ that belong to a finite alphabet $\mathcal{X} = \{\xi_1, \xi_2, \ldots, \xi_{d_X} \}$. The variables are organized here on a plane (as an image) because they will compose the layers of a multi-layer architecture. The $N \cdot M$ variables code multiple  discrete labels, that  in the application that follows take  values in the same alphabet, but they could easily have different cardinalities if we need to fuse  information coming from different sources (the combination of the heterogeneous variables is one of the most powerful peculiarities of the FGrn paradigm). 
Generally the complexity of the whole system increases with the cardinality of the  alphabets. 

The $N \cdot M$ bottom variables are connected to one Hidden (Latent) Variable $S$, that belongs to the finite alphabet $\mathcal{S} = \{\sigma_1, \sigma_2, \ldots, \sigma_{d_S} \}$. The replicator block of Figure \ref{fig:intro}(b)  is drawn here as a box as it will be a patch of the upper plane of each layer in our architecture.  Each connection to the bottom layer is a SISO block that represents the $d_S \times d_X$ row-stochastic probability matrix
$$
P(X[n,m]|S) = 
\begin{bmatrix} 
P(X[n,m] = \xi_1 | S = \sigma_1) \ \hdots \ P(X[n,m] = \xi_{d_X} | S = \sigma_1) \\
P(X[n,m] = \xi_1 | S = \sigma_2) \ \hdots \ P(X[n,m] = \xi_{d_X} | S = \sigma_2) \\
\vdots \\
P(X[n,m] = \xi_1 | S = \sigma_{d_S}) \ \hdots \ P(X[n,m] = \xi_{d_X} | S = \sigma_{d_S}) \\
\end{bmatrix}
$$
The  system is drawn as a generative model with the arrows pointing down assuming that the source is variable $S$ and the bottom variables are its children. This architecture can be seen also as  a Mixture of Categorical Distributions \citep{Koller2009}. Each element of the alphabet $\mathcal{S}$ represents a "Bayesian cluster" for the $N \cdot M$ dimensional stochastic image, ${\bf X} = [X[n,m]]_{n=1:N}^{m=1:M}$ (similar to the Naive Bayes classifier \citep{Barber2012}). 
Essentially each bottom variable is independent from the others given the Hidden Variable \citep{Koller2009}. One way to visualize the model is to imagine drawing a sample: for each data point we draw a cluster index $s \in \mathcal{S}=\{\sigma_1, \sigma_2, \ldots, \sigma_{d_S} \}$ according to the prior  distribution $\pi_S$. Then for each $n=1:N$, $m=1:M$, we draw  $x[n,m] \in \{\xi_1, \xi_2, \ldots, \xi_{d_X} \}$ according to $P(X[n,m]|S=s)$.

We can perform exact inference simply by letting the messages propagate and collecting the results. Information can be injected at any node and inference can be obtained for each variable using the usual sum-product rules. For each SISO block of Figure \ref{fig:lat1} the incoming messages ($b_X$ and $f_S$) and the outgoing messages ($f_X$ and $b_S$) flow simultaneously following the rules outlined in the previous section (sum rule). In the replicator block, incoming messages from all directions are combined with product rule to produce outgoing messages. We can imagine the replicator block as acting like a bus where information is combined and diverted towards the connected branches.   

Handling information in the Bayesian architecture is very flexible since  each  variable $X[n,m]$ corresponds to a pair of messages. The backward message coming from  below is propagated upward towards the latent variable and, through the diverter, towards the sibling branches downwards to the forward messages at the terminations. At the same time the latent variable, fed through its forward message from above, sends information downward through the diverter.

\section{Multi-layer FGrn}
\begin{figure}[!]
\centerline{\includegraphics[width=.8\linewidth]{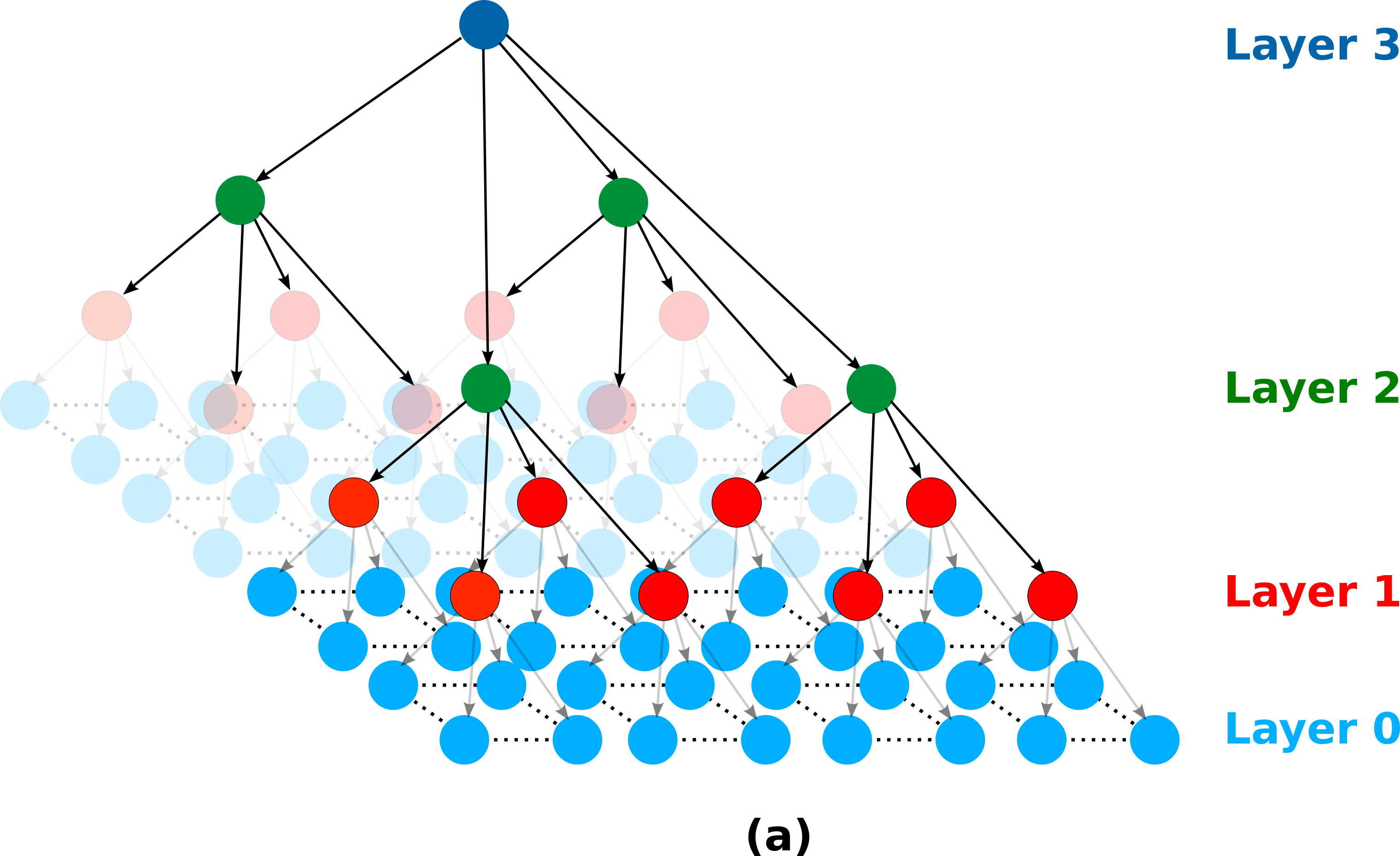}}

\centerline{\includegraphics[width=1\linewidth]{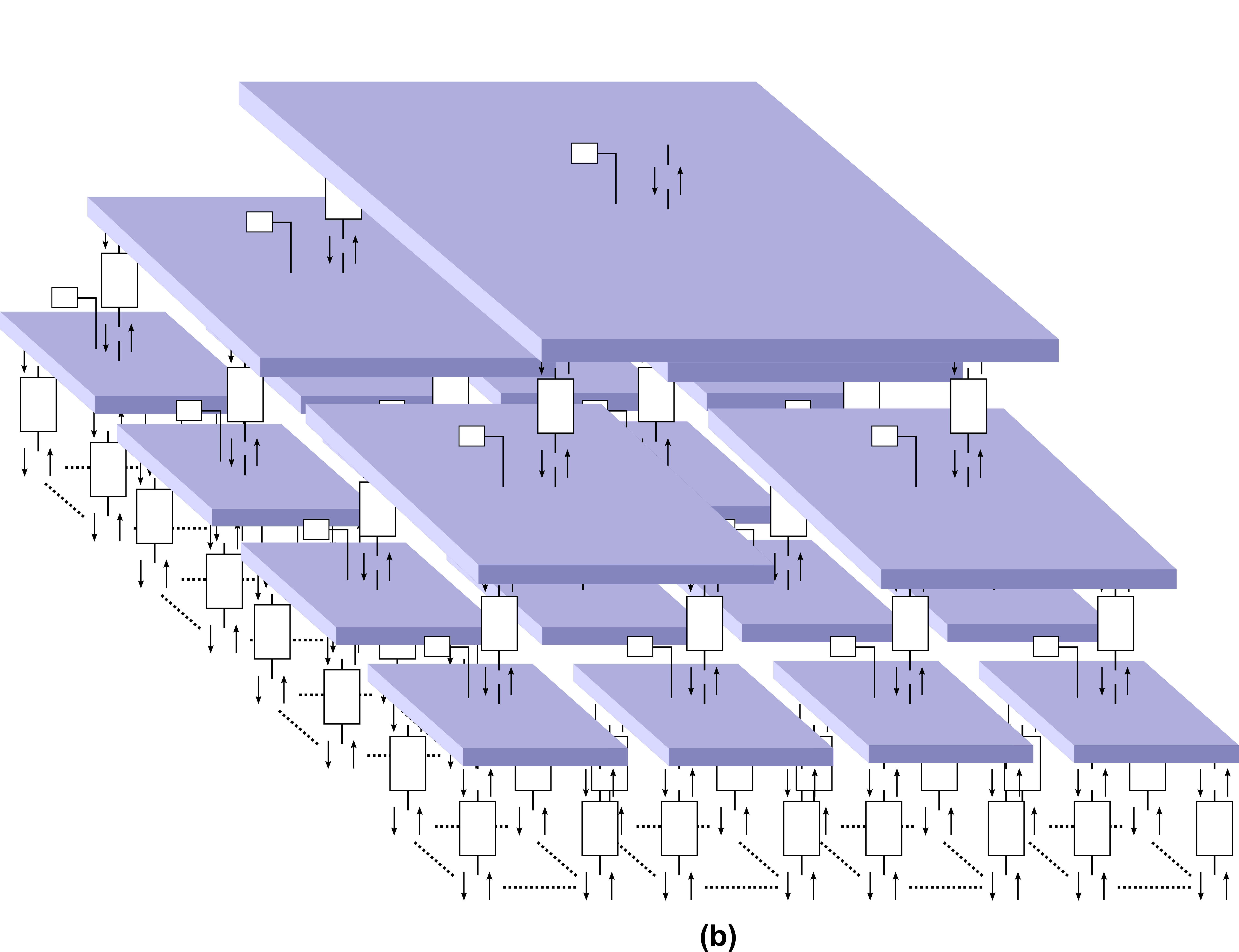}}
\caption{(a) The quadtree architecture; (b) Reduced Normal Factor Graph representation of the quadtree architecture with 4 layers (0-3).}
\label{fig:MFG}
\end{figure}

In this work we build a multilayer structure as in Figure \ref{fig:MFG}(a) on top of a bottom layer of random variables. They can be pixels of an image, a feature map, or more generally a collection of spatially distributed discrete variables. 
In the following we refer to the bottom variables as the {\it Image}. 

The architecture that lays on top of the Image is the  quadtree. In  Figure \ref{fig:MFG} the cyan spheres are the image variables and the other ones (red, green and blue) are the embedding (latent or hidden) variables of the LVM blocks. In Figure \ref{fig:MFG}(b) the same architecture is represented as a FGrn.

A network with $L+1$ levels (Layer $0, \hdots,$Layer $L$) covers a bottom image (Layer $0$) $S_0[n,m]$ $n=1:N \cdot 2^{L-1}$, $m=1:M \cdot 2^{L-1}$, subdivided in $(2^{L-1}) \cdot (2^{L-1})$ image patches of dimension $(N \cdot M)$. 
At  Layer $1$  each patch is managed by one of the latent variables $S_1[n,m]$, $n=1:2^{L-1}$, $m=1:2^{L-1}$ of cardinality of $d_{S_1}$.
At Layer $2$ each latent variable $S_2[n,m]$, $n=1:2^{L-2}$, $m=1:2^{L-2}$ with dimension $d_{S_2}$, is connected to 4 variables of Layer $1$. 
Similarly climbing the tree in quadruples up to the root with variable $S_{L}$. 

Messages travel within each layer (among the subset of LVM variables) and with the layers above and below (within the connected patches and quadruples). The architecture builds a hierarchical Bayesian clustering of the information that is exchanged across different representation scales. 

\subsection{Inference Modes}

If the network parameters have been learned, the system can be used in the following main inference modes: 
\\

\noindent
{\bf Generative}:  A latent variable $S_i[n,m]$ is fixed to a value $\sigma^i_k$, i.e. its forward distribution is a delta $f_{S_i}(s)=\delta(s-\sigma^i_k)$, $\sigma^i_k \in  \mathcal{S}_i=\{\sigma^i_1, \sigma^i_2, \ldots, \sigma^i_{d_{S_i}} \}$. After message propagation downward the forward messages at the terminal variables $S_0[n,m]$ in the cone subtended by $S_i$ reveal the $k$-th "hierarchical conditional distribution"  associated to $S_i$. 
This generation could be done on Layer 1 to check for clusters in the image patches, or at higher layers to visualize  the coding role of the various hierarchical representations.

Of course propagation can be done from a generic node also upward with a backward delta distribution.  The complete upward and downward flow, up to the tree root and down to the other terminations, would reveal  the role of that specific node in the representation memorized in the system (a sort of impulse response).     

\noindent
{\bf Encoding}: The  image ${\bf S_0} = S_0[n,m]$, $n=1:N \cdot 2^{L-1}$, $m=1:M \cdot 2^{L-1}$, is known and the values of the bottom variables are injected in the backward messages as delta distributions. After all messages have been propagated for a number of steps equal to the network diameter (in this case $2 \cdot L+1$), at each hidden variable $S_i[n,m]$, $i=1,\hdots,L$, $n=1:2^{L-i}$, $m=1:2^{L-i}$, we find the contribution of the observations to the posterior for  $S_i[n,m]$. The exact posterior on $S_i[n,m]$ is obtained as the normalized  product of  forward and backward  messages. Each hidden variable represents one of the components of the (soft) code of the image. 
 
\noindent
{\bf Pattern Completion}:  Only some of the bottom variables of ${\bf S}_0$ are known, i.e. their backward messages are deltas. For the unknown variables the backward messages are uniform distributions. In this modality, after at least $2 \cdot L+1$ steps,  the network returns forward messages at the terminal variables that try to complete the pattern ({\em associative recall}, or {\em content-addressed memory}). 

\noindent
{\bf Error Correction}:  Some of the bottom variables of ${\bf S}_0$ are known softly, or wrongly, with smooth distributions, or delta functions respectively. After at least $2 \cdot L+1$ steps of message propagation,  the network produces forward messages at the terminations that attempt to correct the distributions, or reduce the uncertainty. The posterior distributions at the terminal variables $S_0[n,m]$ are obtained as the normalized product of forward and backward messages. 

\smallskip
\noindent
Before propagation, all messages that do not correspond to evidence,  are initialized to uniform distributions. 

\subsection{Learning}
 
The parameters contained in the FG are learned from a training set of  $T$ images $ {\bf S}_0^1, \hdots, {\bf S}_0^{T} $. 

We assume that within each layer the LVM blocks share the same parameters. This is a standard shift-invariance assumption that most deep belief networks make.   

Given a basic patch of dimension $N \cdot M$ pixels and a network with $L+1$ levels (Layer $0, \hdots,$Layer $L$), for Layer $1$ we need to learn $N \cdot M$ matrices $P(S_0[n,m]|S_1)$ (one per pixel) each one having sizes $d_{S_1} \times d_{S_0}$ and the $d_{S_1}$-dimensional prior vector $\Pi_{S_1}$. 

For Layers $2$ to $L$ we need to learn 4 matrices $P(S_{i-1}[n,m]|S_{i})$ having sizes $d_{S_i} \times d_{S_{i-1}}$ and the $d_{S_i}$-dimensional prior vector $\Pi_{S_i}$.

A generic image of the training set is subdivided in $L$-Level patches of dimension $(2^{L-1} \cdot N) \cdot (2^{L-1} \cdot M)$ pixels. 
Each $L$-Level patch is subdivided in $2 \cdot 2$ $(L-1)$-Level patches, $2^{2} \cdot 2^{2}$ $(L-2)$-Level patches and so on until to obtain $(2^{L-1}) \cdot (2^{L-1})$ patches of dimension $(N \cdot M)$ pixels at Layer 1 ($0$-Level and $1$-Level patches are the same).

The examples, subdivided in $1st$-Level patches, are presented to the termination of the LVM in Fig. \ref{fig:lat1} as sharp backward distributions for a fixed number of steps (epochs). All SISO blocks and the source block adapt their parameters  using an iterative Maximum Likelihood Algorithm (\cite{Palmieri2013}) outlined in Section \ref{sec:not}. 

Once the Layer $1$ is learned, the $2nd$-Level patches are used to learn Layer $2$ constructed combining 4 LVMs of Layer $1$ and the process goes on, building deeper and deeper network and considering larger and larger patches.

At the end of the learning phase the matrices are frozen and used in one of the inference modes described before on the same training set to check for accuracy and on a test set to check for generalization.

\begin{figure}[!]
\centerline{\includegraphics[width=1\linewidth]{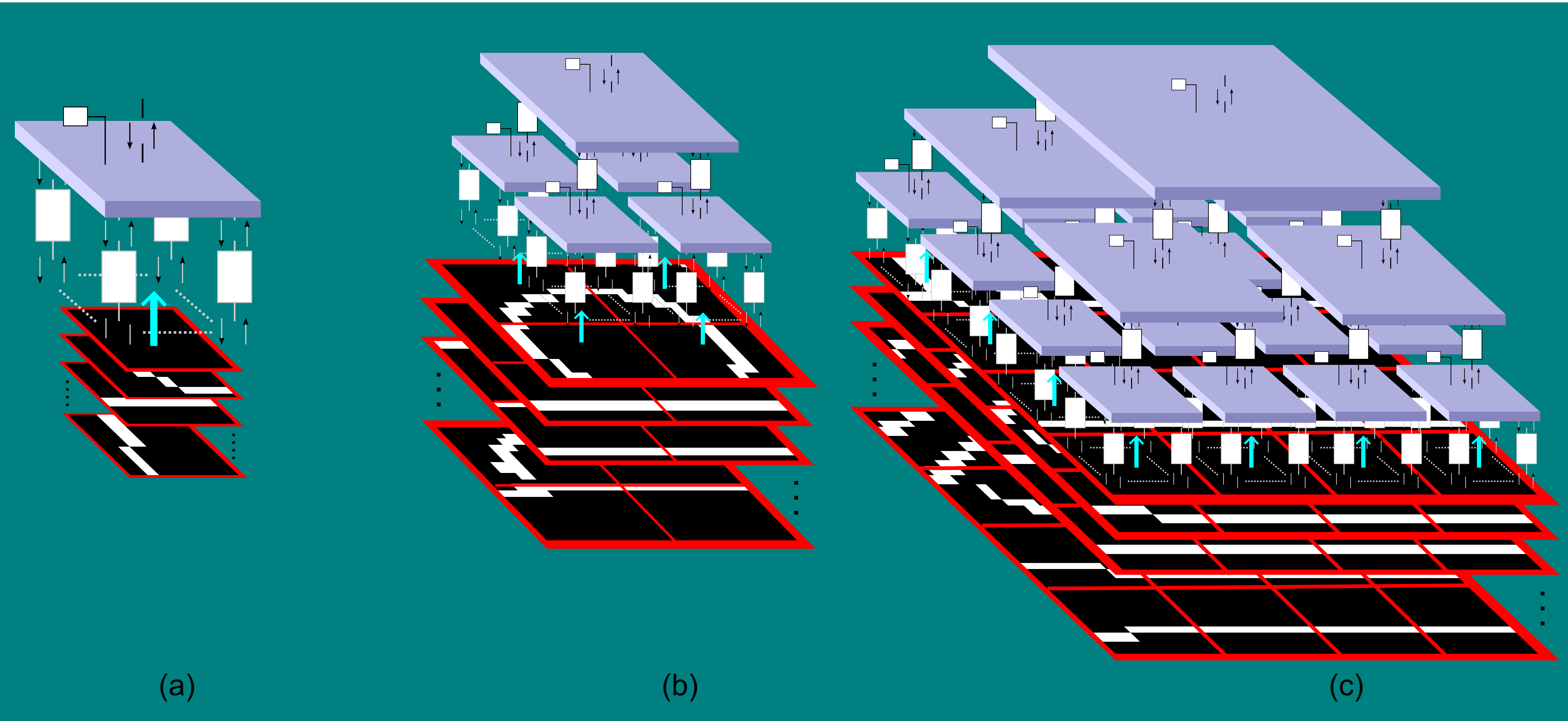}}
\caption{Learning Steps for a 4-Layers Architecture: (a) Learning  Layer $1$; (2) Learning Layer $2$; (3) Learning  Layer $3$}
\label{fig:LearningLayers}
\end{figure}

More specifically, learning is off-line and it is composed by the following steps for an architecture of $L+1$ layers and a basic patch dimension of $N \cdot M$ pixels:
\\
\begin{enumerate}
\item We randomly select $P$ $L$-Level Patches from each image in the Training Set composed by $T$ images. Therefore, in the learning phase, we have $T \cdot P$ $L$-Level patches that are subdivided in Patches of the lower levels until to obtain $1st$-Level Patches;
\item The $T \cdot P \cdot (2^{L-1} \cdot 2^{L-1})$ $1st$-Level Patches of $(N \cdot M)$ pixels are injected at the bottom of the LVM and the parameters are learned (Figure \ref{fig:LearningLayers}(a)); 
\item A new $3$-Layers network (0-2) is built replicating $2 \cdot 2$ times the LVM block learned above and connecting their Hidden Variables with another LVM Block;
\item The  $T \cdot P \cdot (2^{L-2}) \cdot (2^{L-2})$ Patches of $(2 \cdot N) \cdot (2 \cdot M)$ pixels are injected at the bottom of  the new $3$-Layers network and the backward messages at the top of Layer $1$ are used to learn the LVM block at  Layer $2$ (Figure \ref{fig:LearningLayers}(b));
\item A new $4$-Layers network (0-3) is built replicating for $2^2 \cdot 2^2$ times the LVM block learned at step 2 and for $2 \cdot 2$ times the LVM block learned at  step 4 and connecting their Hidden Variables to another LVM Block;
\item The  $T \cdot P \cdot (2^{L-3}) \cdot (2^{L-3})$ Patches of $(2^2 \cdot N) \cdot (2^2 \cdot M)$ pixels are propagated in the Layer $1$ and Layer $2$ and the backward messages at the top of the Layer $2$, are used to learn the LVM block at Layer $3$ (Figure \ref{fig:LearningLayers}(c));
\item The same progression is applied to all the other layers, extending the number of  LVM block replicas to cover the dimension of the current-Level Patch.
\end{enumerate}

\section{Simulations}
\begin{figure}[!]
\centerline{\includegraphics[width=.7\linewidth]{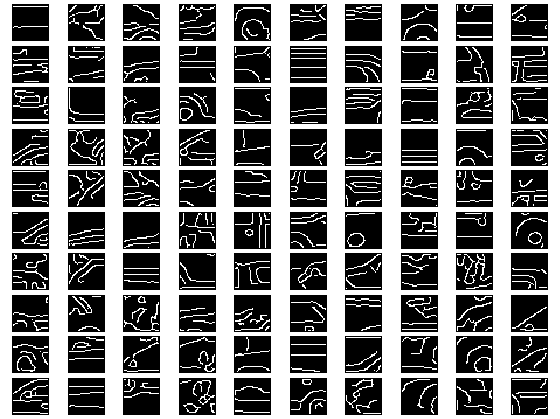}}
\caption{Some images from the Training Set}
\label{fig:TrainingSet}
\end{figure}

In this set of simulations we have taken 50 car images from Caltech101 Dataset. Each image is cropped, filtered with an anisotropic diffusion algorithm (\cite{KovesiMATLABCode}), whitened and finally filtered with a Canny filter in order to obtain images with only the car borders. Our input alphabet is binary ($d_{S_0}=2$). From the 50 filtered images a set of 500 image patches of $32 \cdot 32$ pixels are randomly extracted. A small subset is shown in Fig. \ref{fig:TrainingSet}. 

\subsection{Learning}
The steps for the learning phase are  described in the previous and use the following variables: $P=10$, $T=50$, $N=8$, $M=8$, $L=3$, $d_{S_0} = 2$, $d_{S_1} = 100$, $d_{S_2} = 300$, $d_{S_3} = 300$. 

\subsection{Inference}

\begin{figure}[!]
\centerline{\includegraphics[width=.7\linewidth]{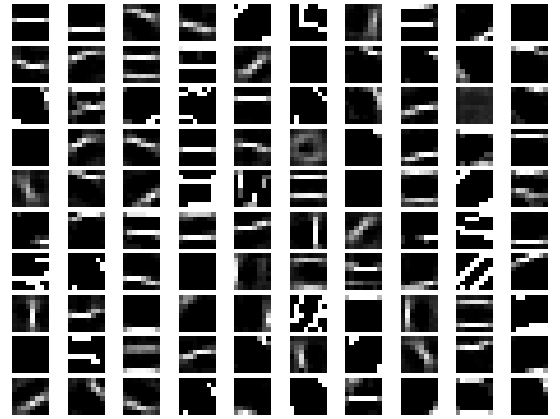}}
\caption{Forward distributions learned at the first level. Dimension of the Embedding Space: 100}
\label{fig:filtersLayer1_100}
\end{figure}

\begin{figure}[!]
\centerline{\includegraphics[width=.7\linewidth]{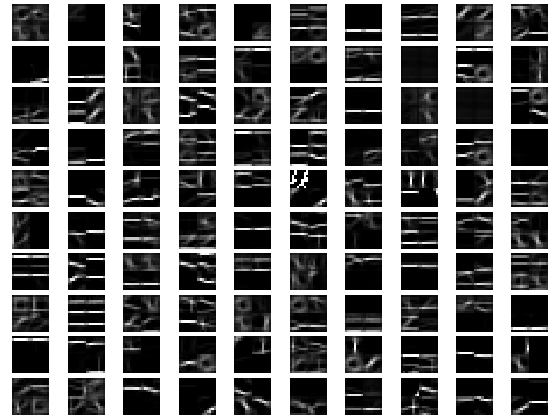}}
\caption{100 of 300 forward distributions learned at the second level. Dimension of the Embedding Space: 300}
\label{fig:filtersLayer2_100}
\end{figure}

\begin{figure}[!]
\centerline{\includegraphics[width=.7\linewidth]{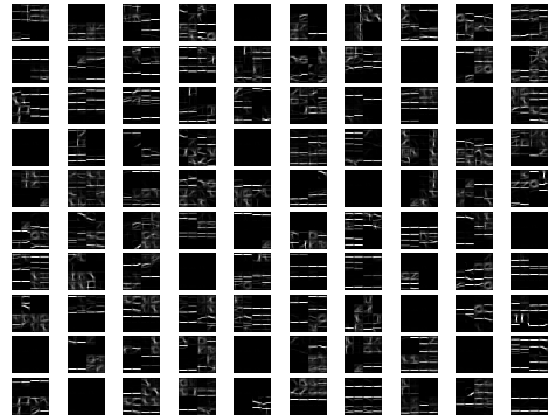}}
\caption{100 of 300 forward distributions learned at the third level. Dimension of the Embedding Space: 300}
\label{fig:filtersLayer3_100}
\end{figure}

Once the  matrices have been learned we  use the network in various inference modes: 

\smallskip
\noindent
{\bf Generative mode}:  We  obtain forward distributions at the bottom terminations  by injecting at the top of the various structures a delta distribution (the images in gray scale show at each pixel the probability on one of the two symbols).  More specifically, for visualizing the conditional distributions corresponding to  Layer $1$ we consider only the Latent Model in Figure \ref{fig:lat1}; for Layer $2$ we consider the $3$-Layers architecture composed by 4 LVM Blocks connected to one LVM Block; for  Layer $3$ we consider the complete architecture. Figures \ref{fig:filtersLayer1_100}, \ref{fig:filtersLayer2_100} and \ref{fig:filtersLayer3_100} show respectively the forward distributions generated injecting deltas at Layers $1$, $2$ and $3$. 

The network has stored quite well the complex structures. The forward distributions from Layer 1 represent simple orientation patterns similar to the ones that the early human visual system responds to. At Layer 3 the distributions reflect the combined representations stored at larger scales. 

\begin{figure}[!]
\centerline{\includegraphics[width=1\linewidth]{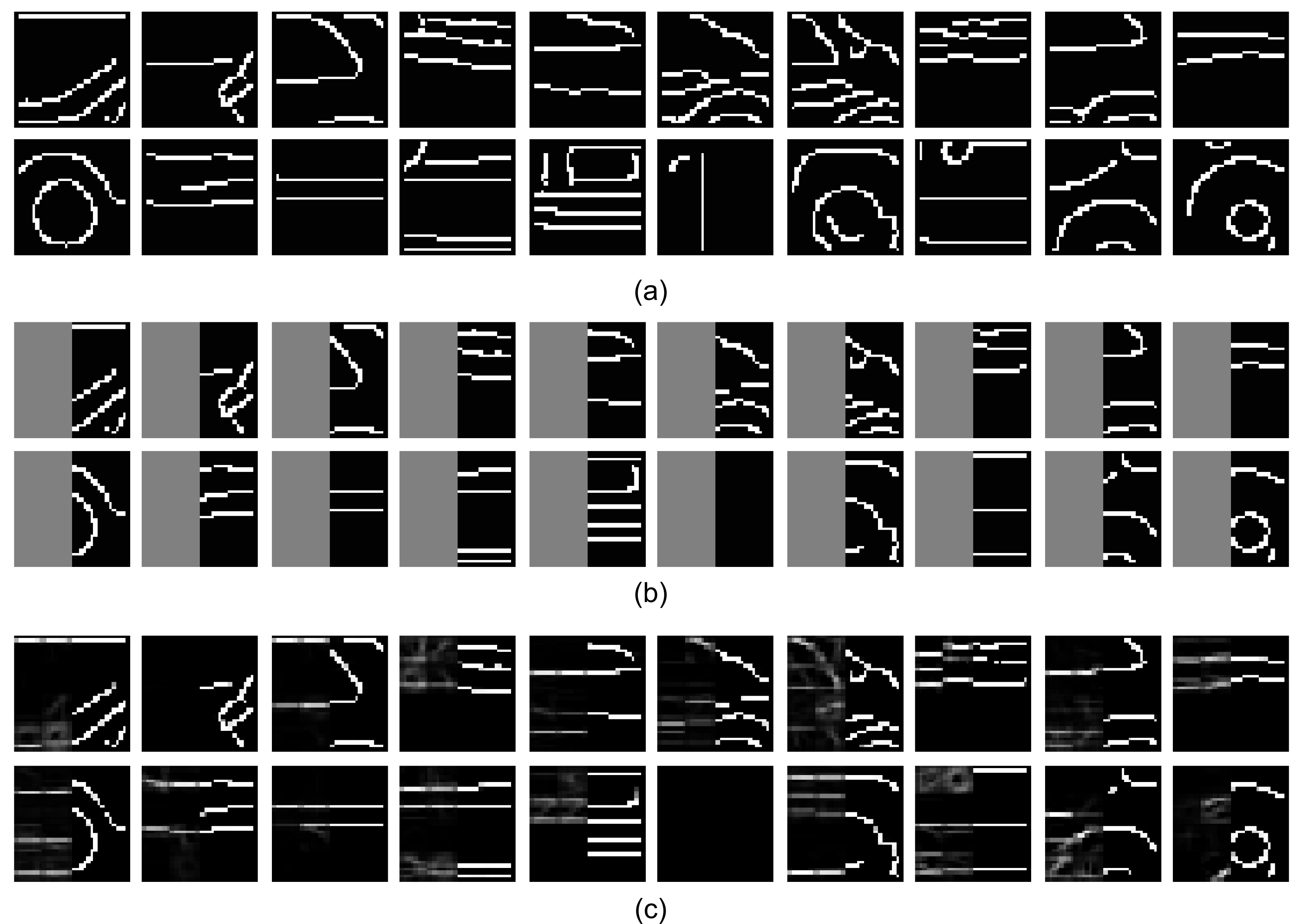}}
\caption{Completion of Patterns obtained from the Training Set. (a) Original image, (b) Image with many erasures (in gray), (c) Image inferred by the network}
\label{fig:ResultTraining}
\end{figure}

\smallskip
\noindent
{\bf Pattern Completion}: In these experiments we have used the architecture as an associative memory that is queried with incomplete patterns and responds with the information  stored during the learning phase. 
Figure \ref{fig:ResultTraining}(a) shows 20 patches extracted from Training Set. Before the injection at the bottom of the network, a considerable amount of pixels (16 patches on a total of 32 patches) has been erased (Figure \ref{fig:ResultTraining}(b)), i.e. the delta backward distribution is replaced with an uniform distribution. Figure \ref{fig:ResultTraining}(c) shows the forward distributions for the same images after message propagation. The network is able to resolve quite well most of the uncertainties using the stored information. 

\begin{figure}[!]
\centerline{\includegraphics[width=1\linewidth]{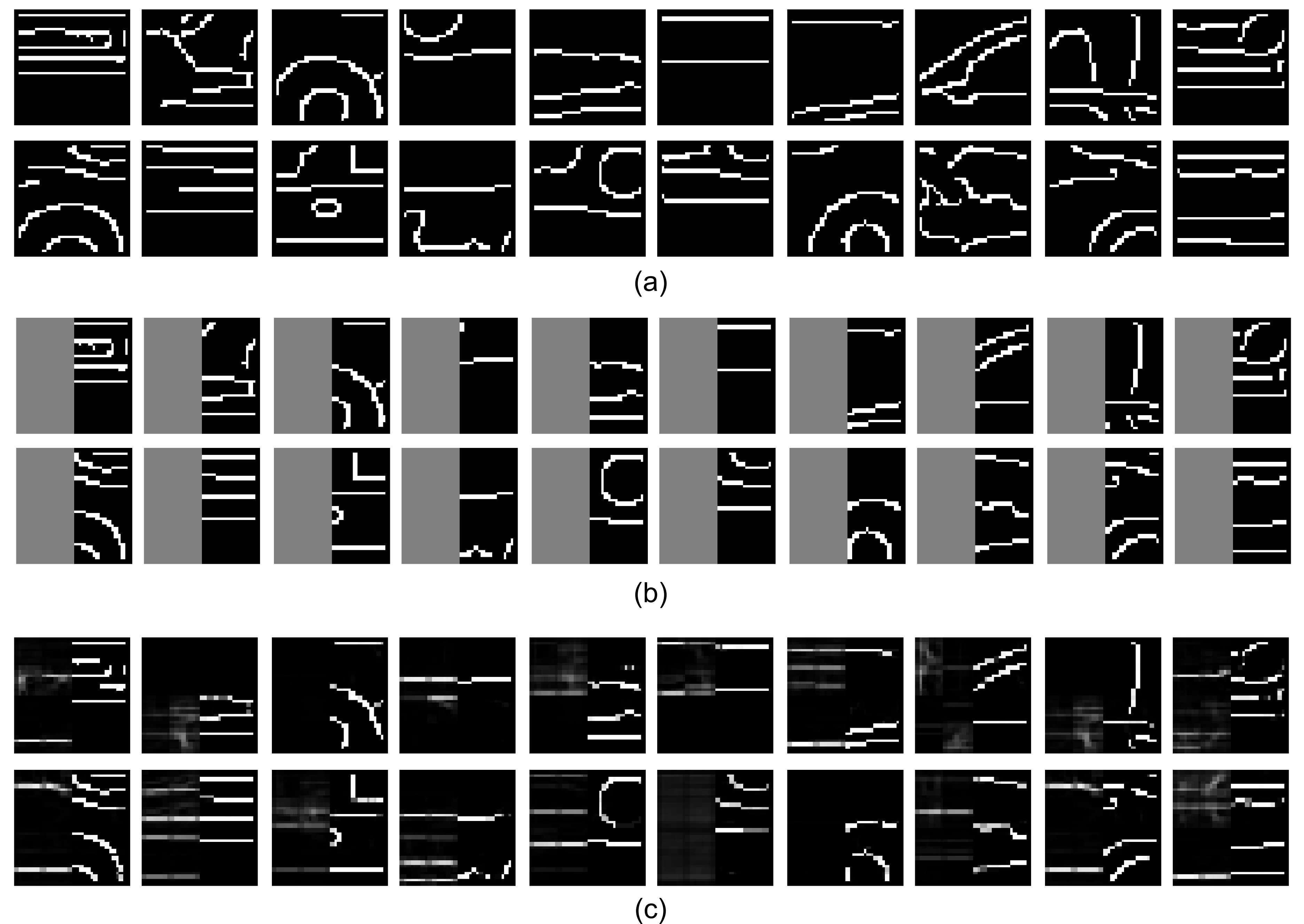}}
\caption{Completion of Patterns obtained from the Test Set. (a) Original image, (b) Image with some erasures (in gray), (c) Image inferred by the network.}
\label{fig:ResultTest}
\end{figure}

\noindent
The same experiment of Pattern Completion has been repeated with  patches extracted from the Test Set (patterns that the network has never seen before). The result is obviously worse as shown in Figure \ref{fig:ResultTest}, but it's worthy to note that the network  succeeds quite well in completing most of the shapes in applying the learned knowledge (test for generalization). Cross-validation can be easily applied to determine the most appropriate embedding spaces sizes for best generalization.  

\section{Discussion and Conclusions}
\label{sec:conc}

From the results presented above we have demonstrated that the paradigm of the FGrn can be successfully applied to build deep architectures. The layers retain the information about the clusters contained in the data and build a hierarchical internal representation. Each layer successfully learns to compose the objects made available from the lower layers. 

We have chosen the border of car images  extracted from  Caltech101 because we wanted to see if the paradigm was suitable for patching together the salient structures of an object. Other experiments have been performed on characters and different patterns revealing very similar results.  

We believe that the FGrn paradigm constitutes a promising addition to the various proposals for deep networks that are appearing in the literature. It can provide great  flexibility and modularity. The network can be easily extended by introducing new and different (in cardinality and in type) variables.  Prior knowledge and supervised information can be inserted at any of the scales: new ``label variables" can be added in one or more of the diverter junctions and let learning take care of  parameter adaptation. Results of these mixed supervised/unsupervised architectures are under way and will be reported elsewhere.

The computational complexity issue that clearly emerges from this paradigm, specially when the embedding variables have large dimensionality and when image pixels are non binary,  is being exploited for parallel implementations. Since both belief propagation and learning are totally local, they can be implemented with distributed hardware or parallelized processes. Some studies have been carried out  for other deep network frameworks \citep{Liang2009}, \citep{Silberstein2008}, \citep{Nam2012}) and we are confident that similarly the FGrn paradigm may present new interesting opportunities to approach some of the most challenging tasks in computer vision.

\bibliography{ProbPropBib}

\end{document}